\documentclass[11pt]{article}
\usepackage{booktabs,threeparttable}
\usepackage{array}
\newcolumntype{C}[1]{>{\centering\arraybackslash}p{#1}}
\usepackage{amsmath}
\usepackage{times}
\usepackage{booktabs}
\usepackage{siunitx}
\usepackage{graphicx}
\usepackage{color}
\usepackage{multirow}
\usepackage[authoryear]{natbib}
\usepackage{rotating}
\usepackage{bbm}
\usepackage{latexsym}
\usepackage{doi}
\usepackage{hyperref}

\providecommand\doi[1]{\href{https://doi.org/#1}{\url{#1}}}
\usepackage{amsthm}
\usepackage{amsfonts}
\usepackage{amssymb}
\usepackage{bbm}

\usepackage[ruled,vlined]{algorithm2e}

\newtheorem*{theorem*}{Theorem}

\newcommand{\captionfonts}{\normalsize}

\makeatletter  
\long\def\@makecaption#1#2{%
  \vskip\abovecaptionskip
  \sbox\@tempboxa{{\captionfonts #1: #2}}%
  \ifdim \wd\@tempboxa >\hsize
    {\captionfonts #1: #2\par}
  \else
    \hbox to\hsize{\hfil\box\@tempboxa\hfil}%
  \fi
  \vskip\belowcaptionskip}
\makeatother

\begin{document}

\begin{center}
{\LARGE Fast, close, non-singular and property-preserving
approximations of entropic measures}\\
{\bf \large Illia Horenko$^{\displaystyle 1}$, Davide Bassetti$^{\displaystyle 1}$, Luk\'{a}\v{s} Posp\'{i}\v{s}il$^{\displaystyle 2}$}\\
{$^{\displaystyle 1}$Chair for Mathematics of AI, Faculty of Mathematics, 
  Rheinland-Pf\"{a}lzische Technische Universit\"{a}t Kaiserslautern Landau, Kaiserslautern, Germany, \\
 $^{\displaystyle 2}$Department of Mathematics,
  Faculty of Civil Engineering,
  VSB - Technical University of Ostrava, Ostrava, Czech Republic
  }\\
\end{center}

\markboth{}{NC instructions}
\ \vspace{10mm}\\
{\bf Abstract: }
Entropic measures like Shannon entropy (SE), its quantum mechanical analogue von Neumann entropy, and Kullback-Leibler divergence (KL) are key components in many tools used in physics, information theory, machine learning (ML) and quantum computing. Besides of the significant amounts of SE and KL computations required in these fields, the singularity of their gradients near zero  is one of the central mathematical reason inducing the high cost,  frequently low robustness and slow convergence of computational tools that rely on these concepts. 
Here we propose the Fast Entropic Approximations (FEA) – non-singular rational approximations of SE and symmetrized KL, that preserve their main mathematical properties and achieve a mean absolute errors of around $10^{-3}$ ($10-20$ times better than comparable state-of-the-art computational approximations).
We show that FEA allows up to around $24$ times faster computation of SE and up to $37$ times faster computation of symmetrized KL: it requires only $5$ to $7$ elementary computational operations, as compared to the tens of elementary operations behind SE and KL evaluations based on approximate logarithm schemes with table look-ups, bitshifts, or series approximations.
On a set of common benchmarks for the feature selection problem in machine learning, we show that the combined effect of fewer elementary operations, low approximation error, preservation of main mathematical properties, and non-singular gradients allows much faster training of significantly-better models. We demonstrate that FEA enables ML feature extraction that is three orders of magnitude faster, and better in quality then the very popular LASSO feature extraction.
 
\newpage
Shannon's entropy (SE) and Kullback-Leibler divergence (KL) are key theoretical and computational components in many areas of science, ranging from physics (like in Ising and Landau-Ginzburg models), to information theory, artificial intelligence (e.g., in Boltzmann machines, Hopfield networks and entropic learning methods) and quantum computing \cite{bengio14,kl14,jiang18,entropy22,ali23}. 
The race for the most efficient SE and KL computation algorithms started almost $80$ years ago, and has led to the development of a multitude of approaches, primarily driven  by the need to (i) preserve the precision of the computation, and to (ii) reduce the number of elementary processor operations to enhance computational speed \cite{shannon48,mitchell62,mitchell13,mitchell17}.
Many designs of tools for SE and KL computation are based on a piecewise-linear logarithm approximation algorithm developed by Mitchell in 1962 \cite{mitchell62,mitchell13,mitchell17}, that requires  $10-12$ elementary operations for a single computation of SE and 20-24 elementary operations for a KL computation. Recently, these approximate algorithms were deployed to speed-up the training of neuronal networks and to make them more energy-efficient: since applying the logarithm transformation allows transforming computationally more expensive and polynomially scaling multiplication and division operations \cite{karatsuba95} to linearly-scaling additions and subtractions \cite{kim18,ansari20,li24}.   However, these algorithms build up on the series approximations like $\log(1+x)\approx x-0.5x^2$, that are accurate for $x\geq1$ - so applying them to the probabilistic arguments $0\leq x\leq1$ requires an additional transformation $1/x$, and results in quite crude approximations with average absolute errors around $0.02$.  Moreover, as was poetically stated in the Section 6.11 about the SE computation  algorithms in the  last edition of the ``Numerical Recipes'': ``... the lurking logarithmic singularity at $x=0$ causes difficulties for many methods that you might try'' \cite{numrec07}. But operating with $x$ being equal or close to zero means finding entropy-minimal and sparse solutions for a broad class of learning problems, like, for example, for model reduction  \cite{tibshirani96}, entropic feature selection  \cite{horenko_pnas_22,horenko23},  and in learning sparse and energy-efficient neuronal networks \cite{yang17,horenko25}.  

To mitigate these problems, in the following Sec. \ref{sec_SE} and \ref{sec_KL} we propose Fast Entropic Approximations (FEA) of SE and symmetrized KL, and show that they provide very close, cheap, and nonsingular approximations that preserve the main mathematical properties of entropic measures. In the Sec. \ref{sec_FS} we will illustrate that the cumulative effect of these FEA properties allows several-orders-of-magnitude boost in performance of the feature selection algorithms from Machine Learning (ML).

\section{Approximation of the Shannon Entropy}
\label{sec_SE}
For a probability distribution $\mathbf{x} = (x_1,\dots,x_N)$ (with all $x_i$ being larger or equal to zero and summing-up to one), Shannon entropy is defined as $\mathcal{H}^{\mathrm{S}}(\mathbf{x})=\sum_{i=1}^N h^{\mathrm{S}}(x_i)$, where $h^{\mathrm{S}}(x) = -x\log(x)$.  

We propose a very simple and compact Fast Entropic Approximation (FEA) $\mathcal{H}^{\mathrm{FEA}}(\mathbf{x})$ of the Shannon Entropy components $h^{\mathrm{FEA}}(x) \approx h^{\mathrm{S}}(x)$:
\begin{eqnarray}
\label{eq:FEA}
\mathcal{H}^{\mathrm{FEA}}(\mathbf{x}) = \sum_{i=1}^N h^{\mathrm{FEA}}(x_i), ~ h^{\mathrm{FEA}}(x) = x\left(\frac{0.6648}{x+0.2086}-0.5754x\right)+0.0206.
\end{eqnarray}

This approximation and its gradient are non-singular, when the arguments $x$ are the probabilities from the interval $\left[0,1\right]$ (see Fig.~1A,1B,1C). 
FEA defined in (\ref{eq:FEA}) approximates SE with a standard natural logarithm, and can easily be re-parameterised to any log-base without additional elementary operations. 
It is straightforward to validate that: 

(i) the mean absolute error of $h^{\mathrm{FEA}}(x)$ with respect to the standard SE $h^{\mathrm{S}}(x)$ on interval $\left[0,1\right]$ is $0.001$ (around 20 times smaller than the mean absolute error $0.0195$ obtained numerically on $\left[0,1\right]$ for the entropy-term approximation induced by the currently-available fastest logarithmic approximations based on Mitchell's algorithm described in \cite{mitchell62,mitchell13,mitchell17}),  
 (ii) that its derivative $(h^{\mathrm{FEA}})'(x)$ is Lipschitz-continuous on the closed interval $\left[0,1\right]$, with the Lipschitz-constant $L\leq31.616$, and (iii) that evaluation of $h^{\mathrm{FEA}}(x)$ requires no more than $6$ elementary operations. 
 
 Moreover, when used in machine learning and AI applications illustrated below, the constant additive term $+0.0206$ can be omitted in optimization problems, thus reducing the required number of elementary operations to just $5$. Lipschitz-continuity of the gradient (point (ii) above), with a relatively-small Lipschitz-constant $L$, is very important for convergence and cost of learning algorithms \cite{sgd19}. As will also be illustrated below: it allows applying  methods like Spectral Projected Gradient (SPG) and Newtons method in AI optimisation \cite{BirMarRayJO-2000,BirMarRayJOSS-2014,nocedal2006numerical,sgd19}, with much faster convergence and much less computational cost then it is possible within the standard AI tools. 
 
 Setting $n=2$ and getting use of the fact that $x_2=1-x_1$, it is straightforward to prove that for $x_1\in[0,1]$, the second derivative of $\mathcal{H}^{\mathrm{FEA}}$ wrt. $x_1$ is negative definite and its gradient is a third-degree polynomial with the single real root $x_1=0.5$. Hence,  for $n=2$, $\mathcal{H}^{\mathrm{FEA}}$ from (\ref{eq:FEA}) preserves the two of the most central mathematical properties of the Shannon Entropy: it attains minima for deterministic probability measures (with $x$ taking the values only zero or one), and has a unique maximum for the uniform probability measure.

\begin{figure}
\centering
\includegraphics[width=\linewidth]{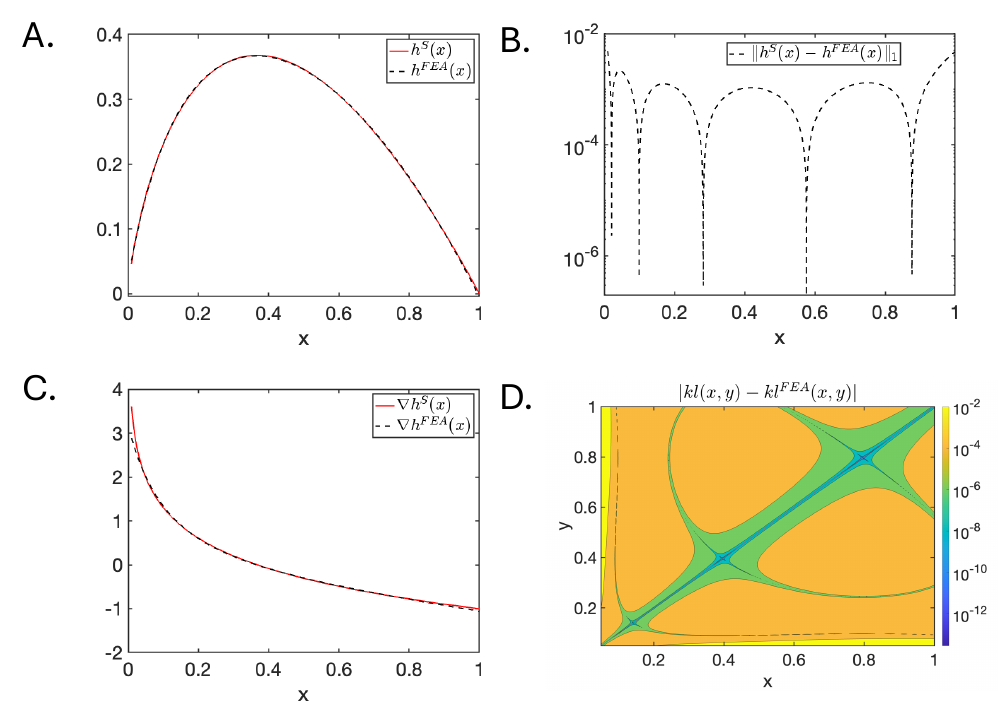}
\caption{\emph{Fast Entropic Approximation (FEA) of Shannon Entropy (SE) (A), absolute approximation errors (B),  comparison of SE and FEA gradients (C), and (D) absolute errors for the FEA of symmetrized Kullback-Leibler divergence.}}\label{fig:1}
\end{figure}

 Next, on several most popular programming language platforms like Python, Julia and Matlab, we compare the  times required to compute the exact values of SE (with the hardwired language-specific functions) to the computational times obtained with FEA, when used in the same language and on the same hardware. These results are summarised in the Table 1 and Fig.~2, revealing up to 24-fold  speed-up for FEA, combined with a very low $1\cdot10^{-3}$ mean absolute approximation error.

\begin{figure}
\centering
\includegraphics[width=\linewidth]{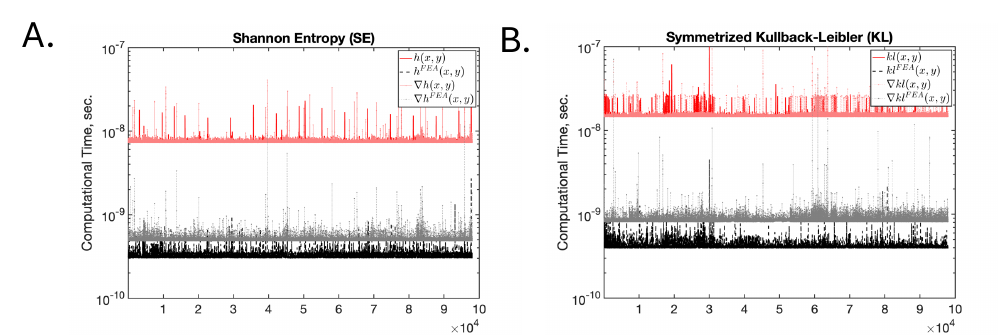}
\caption{\emph{Computational speed-up due to the Fast Entropic Approximation (FEA): (A) for computation of Shannon Entropy and its gradient, and (B) for computation of the symmetrized Kullback-Leibler divergence and its gradient.}}\label{fig:1_5}
\end{figure}

\begin{table*}
   \centering
\caption{\emph{Comparison of computational costs (measured in CPU seconds on a single M1 Apple processor) for the standard Shannon Entropy and its approximation with FEA(\ref{eq:FEA}), with respect to the mean computation times,  mean absolute  error (MAE), and the speed-up factor.  In order to guarantee the fairness of performed  comparisons, we  performed them without parallelization and with only 1 BLAS thread allowed for each computation. Shannon Entropy was computed as} `x.*log(x)' \emph{in MATLAB, as} `x*np.log(x)' \emph{in Python, and as }`@. x*log(x)`\emph{ in Julia.}}
\vspace{0.2cm}
{\tiny
 \begin{tabular}{|p{2.5cm}||c||c|c|}
\hline
\multirow{2}{*}{\parbox{1.5cm}{\textbf{Programming \ language}}} & \multicolumn{1}{c||}{\textbf{Standard computation, seconds}} & \multicolumn{2}{c|}{\textbf{FEA computation, seconds}} \\
\cline{2-4}
& mean time & mean time &  speed-up factor \\ \hline
MATLAB &$7.39\cdot10^{-9}$&$3.1\cdot10^{-10}$&23.8\\ \hline
Python &$4.66\cdot10^{-9}$ &$1.21\cdot10^{-9}$ &3.9 \\ \hline
Python (jit)&$5.09\cdot10^{-9}$ &$6.50\cdot10^{-10}$ &7.91 \\ \hline
Julia &$5.50\cdot10^{-9}$&$7.44\cdot10^{-10}$&9.66 \\ \hline
\end{tabular}
}
 \end{table*}
\normalsize

\section{Approximation of the symmetrized Kullback-Leibler divergence}
\label{sec_KL}
Symmetrized Kullback-Leibler divergence (KL) (sometimes also called Jensen-Shannon divergence) deploys the concept of Shannon entropy to quantify the distance between two N-dimensional probability distributions  $\mathbf{x} = (x_1,\dots,x_N)$ and $\mathbf{y} = (y_1,\dots,y_N)$. It is is defined as $\mathbf{KL}^{\mathrm{sym}}(\mathbf{x}, \mathbf{y})=\sum_{i=1}^N kl(x_i,y_i)$, where $kl^{\mathrm{sym}}(x,y) = -0.5(x-y)\left(\log(y)-\log(x)\right)$. Although it is not a proper distance measure in a strict mathematical sense, it has a set of important mathematical properties that make it close to proper metric, making it very useful in many practical applications: (i) it is nonnegative, (ii) it is symmetric, (iii) it is convex, and (iii) it can be interpreted as a capacity of a noisy information channel with two inputs giving the output distributions $\mathbf{x}$ and $\mathbf{y}$. Moreover, being a member of the family of f-divergences, it is locally proportional to the Fisher information metric \cite{bengio14,nielsen19}. It averages the expectations over log-likelihood ratios of the two distributions, and, according to the Neyman–Pearson lemma, the most powerful test for distinguishing between two distributions is based on the likelihood ratio \cite{pearson33}. 

Deploying the FEA approximation (\ref{eq:FEA}), symmetrically regrouping the terms and re-parameterizing, we obtain the following Fast Entropic Approximation (FEA) $kl^{\mathrm{FEA}}(\mathbf{x},\mathbf{y})$ of the symmetrized KL components $kl^{\mathrm{FEA}}(x,y) \approx kl^{\mathrm{sym}}(x,y)$:
\begin{eqnarray}
\label{eq:FEA-KL}
\mathcal{KL}^{\mathrm{FEA}}(\mathbf{x},\mathbf{y}) = \sum_{i=1}^N kl^{\mathrm{FEA}}(x_i,y_i), ~ kl^{\mathrm{FEA}}(x,y) =\frac{0.3011\left(x-y\right)^2}{(x+0.1636)(y+0.1636)+0.3011}.
\end{eqnarray}

As in the case of the Shannon Entropy approximation with FEA  (\ref{eq:FEA}), it is straightforward to validate that this approximation and its gradient are non-singular, when the arguments $x$ and $y$ are the probabilities from the interval $\left[0,1\right]$. Approximation defined in (\ref{eq:FEA-KL}) approximates KL with a standard natural logarithm, and can easily be re-parameterised to any log-base without additional elementary operations. 
The mean absolute error of $kl^{\mathrm{FEA}}(x,y)$ with respect to the standard  $kl^{\mathrm{sym}}(x)$ on intervals $\left[0,1\right]$ is $0.0019$ (see Fig.~1D), it is around 10 times smaller then the currently-available fastest approximations based on Mitchell's algorithm \cite{mitchell62,mitchell13,mitchell17}). Evaluation of $kl^{\mathrm{FEA}}(x,y)$ requires no more then $7$ elementary operations. As can be seen from the Table 2 and Fig.~2, this significant reduction in the number of elementary operations results in up to 37-fold computational speed-up.
\begin{table*}[h]
   \centering
\caption{\emph{Comparison of computational costs (measured in CPU seconds on a single M1 Apple processor) for the symmetrized Kullback-Leibler divergence and its approximation with FEA(\ref{eq:FEA-KL}),  with respect to the mean computation times,  mean absolute  error (MAE), and the speed-up factor. In order to guarantee the fairness of performed  comparisons, we  performed them without parallelization and with only 1 BLAS thread allowed for each computation. Kullback-Leibler divergence was computed as  }`-0.5*(x-y) .* log(y./x)`\emph{ in MATLAB, as  }`-0.5*(x-y) * np.log(y/x)`\emph{ in Python, and as }`@. -0.5 * (x-y) *(log(y)-log(x))`\emph{ in Julia.}}
\vspace{0.2cm}
{\tiny
 \begin{tabular}{|p{2.5cm}||c||c|c|}
\hline
\multirow{2}{*}{\parbox{1.5cm}{\textbf{Programming \ language}}} & \multicolumn{1}{c||}{\textbf{Standard computation, seconds}} & \multicolumn{2}{c|}{\textbf{FEA computation, seconds}} \\
\cline{2-4}
& mean time & mean time &  speed-up factor \\ \hline
MATLAB &$1.5\cdot10^{-8}$&$4.0\cdot10^{-10}$&36.8\\ \hline
Python &$5.86\cdot10^{-9}$&$1.76\cdot10^{-9}$ &3.41 \\ \hline
Python (jit)&$6.72\cdot10^{-9}$ &$9.97\cdot10^{-10}$ &7.20 \\ \hline
Julia &$1.04\cdot10^{-8}$ &$8.12\cdot10^{-10}$ &16.25 \\ \hline
\end{tabular}
}
 \end{table*}
\normalsize

Moreover, it is easy to verify that $\mathcal{KL}^{\mathrm{FEA}}(\mathbf{x},\mathbf{y})$ defined in (\ref{eq:FEA-KL}) preserves the four main mathematical properties of the exact symmetrized KL: (i) $\mathcal{KL}^{\mathrm{FEA}}(\mathbf{x},\mathbf{y})\geq0$ for all probability measures $x$ and $y$ (non-negativity property); (ii)  $\mathcal{KL}^{\mathrm{FEA}}(\mathbf{x},\mathbf{y})=\mathcal{KL}^{\mathrm{FEA}}(\mathbf{y},\mathbf{x})$ (symmetry);  (iii) $\mathcal{KL}^{\mathrm{FEA}}(\mathbf{x},\mathbf{y})=0$ if and only if $x\equiv y$ (zero property), and (iv) that $\mathcal{KL}^{\mathrm{FEA}}(\mathbf{x},\mathbf{y})=0$ is a convex function (convexity property).

\section{Application to feature selection problem in ML}
\label{sec_FS}
As shown recently, incorporating SE minimization into the learning objective in AI and ML problems enables effective probabilistic  distinguishing among relevant and irrelevant features - and can lead to an effective sparsification and feature extraction in learning problems \cite{horenko_pnas_22,horenko23}.
Such entropic feature selection and model sparsification capabilities — incorporated, for example, within the Sparse Probabilistic Approximation for Regression Task Analysis (SPARTAn), which uses SE for feature selection —  was recently demonstrated to allow for compact, high-performing, and energy-efficient alternatives to common more expensive pruning and feature selection techniques like LASSO-regularization \cite{horenko23,horenko25}.

\begin{figure}
\centering
\includegraphics[width=\linewidth]{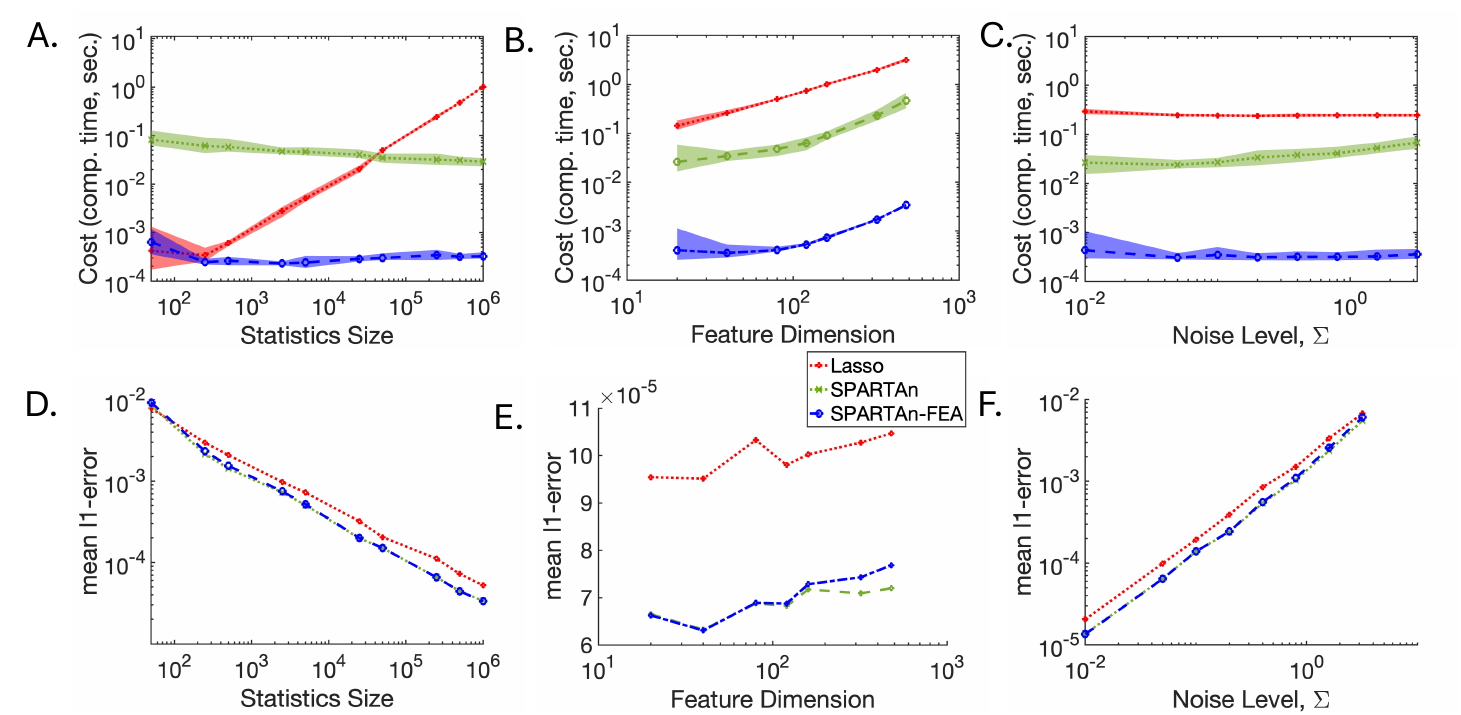}
\caption{\emph{Effect of FEA in ML feature selection: comparisons of SPARTAn (green) and LASSO (red) feature selection algorithms to the SPARTAn that incorporates FEA (SPARTAN-FEA, blue).}}\label{fig:2}
\end{figure}

\begin{figure}[t]
\centering
\includegraphics[width=\linewidth]{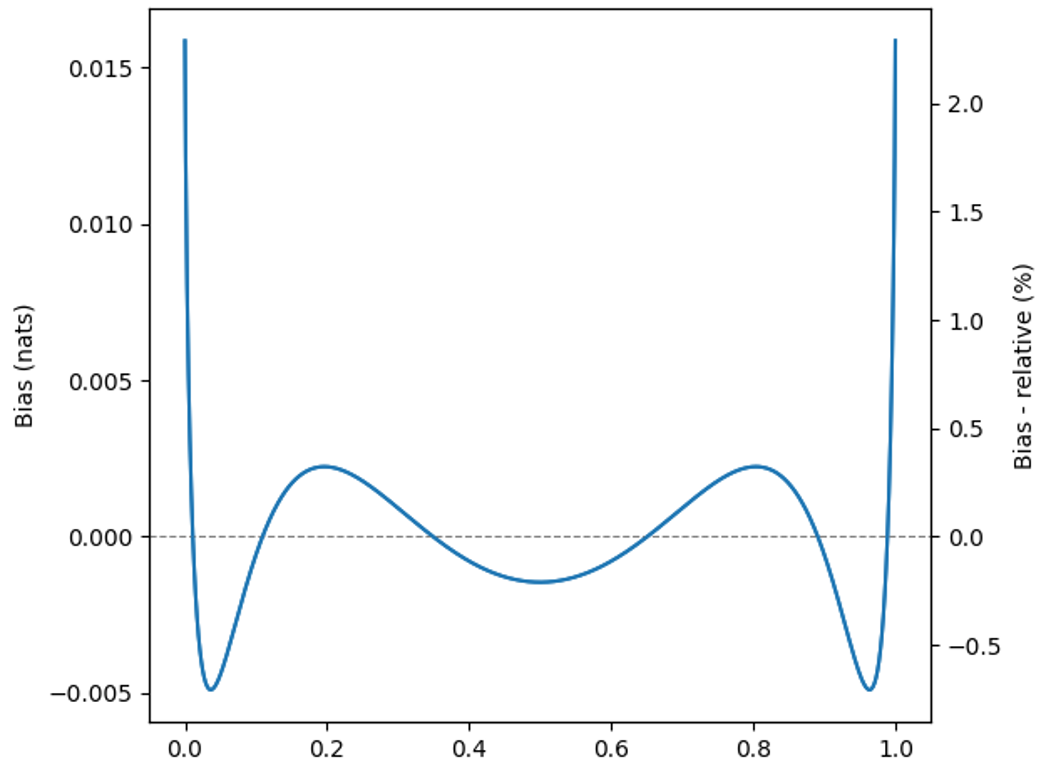}
\caption{\emph{Bias of FEA in the approximate Shannon Entropy computation.}}\label{fig:4}
\end{figure}

To test for potential pitfalls of using FEA to approximate Shannon entropy in this context — and to quantify its benefits — we consider the model reduction problems proposed by R. Tibshirani in his pioneering work on data-driven regression model reduction using the LASSO methodology. The goal is to identify `crisp' and relevant sets of non-zero model parameters, distinguishing them from non-relevant ones that are statistically reduced to zero \cite{tibshirani96}.
This set of test benchmark models entails the randomly generated data of  $N$-dimensional features $\mathbf{F}(t)$ that give rise to the $T$ realisations of the regression $y(t)=\left(\mathbf{x}, \mathbf{F}(t)\right)+\Sigma\cdot\text{noise}(t)$, where $t$ goes from $1$ to the statistics size $T$, $\mathbf{x}$ is a sparse set of fixed and pre-defined regression parameters, and $\left(\cdot,\cdot\right)$ denotes the scalar product. Then, for varying  statistics sizes $T$, feature dimensions $N$, noise levels $\Sigma$ and randomly-generated feature vectors $\mathbf{F}(t)$, as well as with random noise realisations $\text{noise}(t)$, one can create sets of data pairs $\left\{\mathbf{F}(t),y(t)\right\}$ that can be used by different learning methods to infer the (sparse) regression parameters $\mathbf{x}^*$ - and to compare the performance of different learning methods on these data, in terms of the goodness of parameter fit (e.g., with the average $l1$ error norm $\frac{1}{N}\sum_{i=1}^N\left|x^*_i - x_i\right|$) and in terms of the computational cost required for training.

Next, we replace Shannon entropy and its gradient in the SPARTAn algorithm \cite{horenko23} with the FEA approximation (\ref{eq:FEA}) and its gradient. Making use of their Lipschitz continuity, we also substitute the interior-point optimizer — often hindered by the singularity of the SE gradient — with the much faster Spectral Projected Gradient (SPG) optimizer \cite{BirMarRayJO-2000,BirMarRayJOSS-2014}, which was previously inapplicable due to this singularity.
Fig.~1 summarizes results of comparing the new SPARTAn-FEA learner to the SPARTAn \cite{horenko23} and LASSO algorithms for varying statistics sizes, feature dimensions and noise levels. 
As can be seen from Fig.~3D,~3E,~3F, SPARTAN-FEA pertains the goodness of fit of the original SPARTAn, outperforming LASSO in this respect. At the same time, Fig.~3A,~3B,~3C show that  SPARTAN-FEA requires two orders of magnitude less computation then SPARTAn and around thee orders of magnitude less computation then LASSO. 
An implementation of both SE and symmetrized KL divergence in MATLAB is available at \url{https://github.com/EntropicLearning/FEA}.

\section{Discussion}
\label{sec_D}
Shannon entropy (SE) is considered optimal in information theory because it's the \emph{unique non-parametric measure} of uncertainty that satisfies a set of axioms considered necessary for any reasonable measure of uncertainty and information content. However,  singularity of its derivative (and of derivatives of other popular measures based on SE, like KL-divergence) impose considerable difficulties when deploying in learning tools that train with gradient descent and related methods. Standard way of avoiding this problem is in deploying the nonsingular parametric measures, like the Renyi-entropies $\mathcal{H_\alpha}$ (with a tuneable parameter $\alpha$) - but this comes with the cost of adding and tuning this additional hyperparameter, and adding an additional complexity dimension into the problem.    

In this work we have provided an alternative way of addressing this problem, the way that does not require additional hyperparameters. It is based on a very close approximation (Fig.~1A) of the original non-parametric SE by means of the very compact non-parametric Fast Entropic Approximation (\ref{eq:FEA}). Besides of the removing singularity, it was demonstrated to achieve up to 24-fold and 37-fold speed-ups in the approximate SE and KL computations, respectively - while simultaneously improving factor 10-20 the approximation quality as compared to existent algorithms for SE and KL approximation (see Fig~2 and Tables 1 and 2). As was shown above, FEA not only allows obtaining much cheaper and very close non-singular rational approximations, it also preserve the most central mathematical properties of entropic measures- like the extremal properties of SE, as well as the non-negativity, symmetry, convexity and zero-property of symmetrized KL. The word optimality can be also used in relation to FEA, understanding the “optimality” in the sense of l2-approximation: i.e., changing the values of the provided coefficients in the rational approximation that we found, results in increase of the l2-approximation error with respect to the original SE and KL measures. 

The bias introduced by the FEA approximation is systematic and deterministic. 
For a Bernoulli random variable, the bias can be visualized as described in Fig. 4. Its magnitude is smaller than 0.003 nats for a big portion of the interval [0,1], growing to around 0.015 nats at the extremes. The bias is at worst around 2.3\% of the maximal entropy value – and this happens at the ends 0 and 1 of the probability interval. Please note that  it does not change the fact that, for example, FEA remains a concave function, with the two minima at the ends of the interval. Hence, using FEA as a regularisation does not shift the positions of the minima and maxima of the original entropy or KL function – and, during AI training, approaching the points 0 and 1  means approaching the parameter regions where the absolute values of the gradients for both the original and exact Shannon entropy and the KL divergence go to infinity - and where any AI learning algorithm that involves gradients gets a maximal bias and exhibits severe numerical issues anyway. At the same time, around these points 0 and 1 gradients computed from FEA remain well-defined and bounded. Hence, applying FEA actually allows to unbias the numerics of the AI learning algorithms that involve entropy and KL-measures - since, basically all the contemporary numerical training algorithms rely on the computation of gradients of the loss function, and provide systematic bias at the probability interval ends. 

This is confirmed by the results provided in the Fig.~3,  showing that the deployment of FEA in ML and AI algorithms allows a  three orders of magnitude cost reduction when performing feature extraction problems - when compared to the common LASSO-sparsification on common benchmark problems. This opens a possibility for tackling substantially larger feature extraction and learning problems from the natural sciences \cite{horenko_pnas_22,horenko23}, as well as a new opportunity to enhance the pruning and quantization of much large neural networks \cite{yang17,horenko25}, without an increase in computational or hardware requirements.

\section*{Acknowledgement}  This work was funded by the European Commission under Horizons European Programme (Grant Agreement No. 101080756), for the AI4LUNGS consortium, and travel support was provided by MLKL (Machine Learning Consortium in Kaiserslautern).

\bibliographystyle{unsrt}

\end{document}